%% file: main.tex
\documentclass{article}

\usepackage[final]{neurips_wrl2020}

\usepackage[utf8]{inputenc} 
\usepackage[T1]{fontenc}    
\usepackage{hyperref}       
\usepackage{url}            
\usepackage{booktabs}       
\usepackage{amsfonts}       
\usepackage{nicefrac}       
\usepackage{microtype}      

\usepackage{amssymb,graphicx}
\usepackage{subcaption}
\usepackage{wrapfig}

\usepackage{glossaries}

\input{math_commands}

\input{acronyms}

\title{Structured Policy Representation: Imposing Stability in arbitrarily conditioned dynamic systems}

\author{%
  Julen Urain \thanks{Correspondence to \texttt{\{urain, tateo, ren, peters\}@ias.informatik.tu-darmstadt.de}} 

  \And Davide Tateo \footnotemark[1] \And Tianyu Ren \footnotemark[1] \And Jan Peters \footnotemark[1]
  \AND \\
  Intelligent Autonomous Systems, Technische Universität Darmstadt\\

}

\begin{document}

\maketitle

\begin{abstract}
  We present a new family of deep neural network-based dynamic systems. The presented dynamics are globally stable and can be conditioned with an arbitrary context state. We show how these dynamics can be used as structured robot policies. Global stability is one of the most important and straightforward inductive biases as it allows us to impose reasonable behaviors outside the region of the demonstrations.
\end{abstract}

\section{Introduction}

Stability is a major concern when the learned policy is running on a real system, where it is crucial to ensure that the generated trajectories converge to a well-defined attractor e.g., a target position or a limit cycle. In most of the robotics manipulation tasks, the robot will perform a trajectory that could be thought of as a trajectory sampled from a stable dynamic system. This is the case of peg-in-a-hole, pouring, or opening a door.

We considered the problem of representing a family of stable policies. These policies could be applied rather in an Imitation Learning~(\cite{schaal1997learning}) or a Reinforcement Learning~(\cite{peters2008reinforcement}) scenario. Imposing stability in a policy could be beneficial not only for safety issues but also in a learning scenario. By designing inherently stable policies we impose a prior that helps in the generalization, particularly in areas that are far from the previously experienced demonstrations or not explored by the agent.

In our work, we present a new family of globally stable stochastic policies that could represent go-to or cyclic motions. Our proposed policy structure takes advantage of the Conditioned \gls{inn}, to learn rich and highly nonlinear stochastic stable dynamics that could be applied as robot policies. Our work extends \cite{urain2020imitation}, by considering a richer set of \gls{inn}, deeper stability analysis, and adding arbitrary conditioning.

\section{From Policies to Dynamic Systems}

A policy $\pi$, is a function that maps a state $\vs$ to a probability distribution over the action $\va \sim \pi(\va | \vs)$.

In Robotic Manipulation, the action is usually the robot's joint torques($\vtau$), and the state is a combination of joint positions, velocities($\vq, \dot{\vq}$), and some sensor inputs($\vc$). 
As commercial robots often provide access to the dynamic model, we can write mathematically how the joint torques affects the robot's joint state
{\small
\begin{align}
    \vtau + \vtau_{ext} = M(\vq)\ddot{\vq} + C(\vq, \dot{\vq})\dot{\vq} + \vg(\vq),
    \label{eq:dynamics}
\end{align}
}
where $M(\vq)$ is the joint-space inertia matrix, $C(\vq, \dot{\vq})$ is the matrix related with the coriolis and centrifugal forces, $\vg(\vq)$ is the joint-space gravity force vector and $\vtau_{ext}$ is the joint-space external torque forces. Given the robot dynamics are known, we can select a policy that cancels out the robot dynamics and depends exclusively in a certain desired acceleration $\ddot{\vq}_{des}$,~(\cite{4788393})
{\small
\begin{align}
    \vtau = C(\vq, \dot{\vq})\dot{\vq} + \vg(\vq) + M(\vq)\ddot{\vq}_{des}.
    \label{eq:policy}
\end{align}
}
Combining Eq.~\ref{eq:dynamics} and Eq.~\ref{eq:policy},
{\small
\begin{align}
         \ddot{\vq} = \ddot{\vq}_{des} + M^{-1}(\vq)(\vtau_{ext});
\end{align}
}
robot's joint dynamics depends only on our desired acceleration and the external torques. This policy structure is useful for robotic manipulation as we can abstract our policy from torque-space to acceleration-space 
{\small
\begin{align}
    \ddot{\vq}_{des} \sim \pi(\ddot{\vq}_{des}|\dot{\vq}, \vq, \vc).
    \label{eq:dynamic_pol}
\end{align}
}
In this context, \eqref{eq:dynamic_pol} represents a stochastic dynamic system in the robot's joint-space, conditioned on some context information $\vc$. Therefore, we can frame our policy as a stochastic dynamic system and impose properties such as stability on it.   

\section{Stable dynamics with Invertible Neural Networks}

In recent years, \gls{nf}~(\cite{rezende2015variational}) have gained popularity as powerful networks for density estimation, variational inference, and generative modeling. \gls{nf} are explicit likelihood models that construct flexible probability distributions of high-dimensional data composed of a latent distribution from which is easy to sample, $\vz \sim p(\vz)$, usually a normal distribution; and a \gls{inn}, $\vy = f(\vz)$, that maps the latent space $\mZ$ to observation space $\mY$. By the use of change of variable rule, the density on the observation space can be computed in terms of the density in latent space and the invertible transformation's jacobian
{\small
\begin{align}
   p(\vy) = p(\vz) \left|\textrm{det} \frac{\partial f}{\partial \vz} \right|^{-1} 
\end{align}
}
\gls{nf} can be extended to learn dynamic data distributions, by switching the latent static distribution with a dynamic one. Given a latent stochastic dynamic system and an invertible transformation
{\small
\begin{align}
    \vz_{k-1} = f^{-1}(\vy_{k-1}) &  & \vz_k \sim p(\vz_{k}| \vz_{k-1}) & & \vy_{k} = f(\vz_{k}),
    \label{eq:iflow}
\end{align}
}
the stochastic dynamics in $\mY$ can be computed in terms of the stochastic dynamics in $\mZ$
{\small
\begin{align}
\label{eq:dynNF}
    p(\vy_{k}|\vy_{k-1}) = p(\vz_{k}|\vz_{k-1}) \left|\textrm{det} \frac{\partial f}{\partial \vz_{k}} \right|^{-1}.
\end{align}
}
\paragraph{Stability} We have considered the family of stochastic dynamic systems represented with the model proposed in \eqref{eq:iflow}. Given the latent dynamics, $p(\vz_{k}| \vz_{k-1})$, are globally stochastically stable, we prove in the Appendix~\ref{appendix:stability}, that $p(\vy_{k}| \vy_{k-1})$ will also be globally stochastically stable.

\paragraph{Conditioning} As presented in \eqref{eq:dynamic_pol}, a desired robotic policy should be able to consider an arbitrary context state, $\vc$. Anyway, the dynamics we have presented in \eqref{eq:dynNF} can only consider robot's joints. We have switched \gls{inn} with Conditioned \gls{inn} to adapt the dynamics to the context state $\vc$
{\small
\begin{align}
    \label{eq:CdynNF}
    p(\vy_{k}|\vy_{k-1}, \vc_{k}) = p(\vz_{k}|\vz_{k-1})\left|\textrm{det} \frac{\partial f(\vz_k, \vc_{k})}{\partial \vz_{k}} \right|^{-1}.
\end{align}
}
Given that stability is imposed in the structure of the neural network; our model will remain stable for any arbitrary context. Moreover, while most of the previous works (\cite{khansari2011learning, calinon2016tutorial, sindhwani2018learning}), only considered affine context-based transformations(different goal target, different velocities \dots), our model can learn any nonlinear context-based transformation while remaining stable.

\section{Experiments}
We have considered three experiments to validate our model. In the first experiment, we considered a behavioral cloning problem on which a highly nonlinear limit cycle is learned from data. In the second experiment, we wanted to evaluate the performance of our policies when increasing the dimensionality of an arbitrary context. Finally, in the last experiment, we evaluate the performance of our model in a real robot scenario, with switching contexts and external perturbances.

\subsection{Conditioned Vitruvian man}
\begin{figure}
  \centering
  \begin{subfigure}{.74\textwidth}
    \centering
    \includegraphics[width=0.99\textwidth]{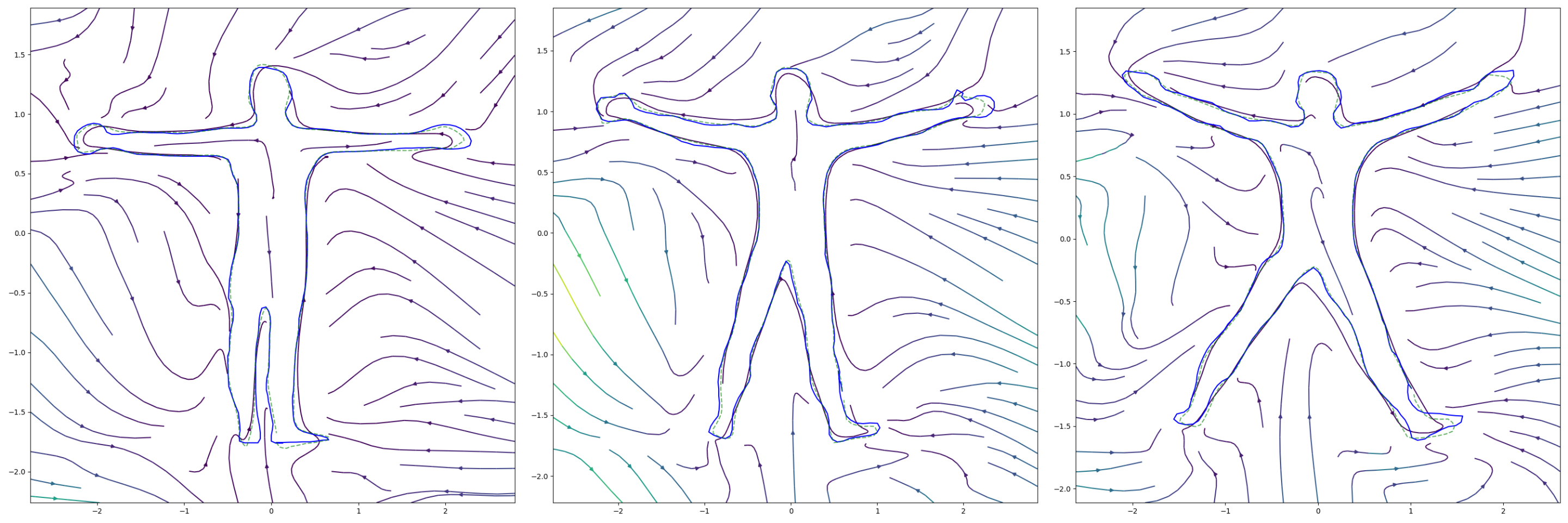}
  \end{subfigure}
  \begin{subfigure}{.25\textwidth}
    \centering
    \includegraphics[width=.9\textwidth]{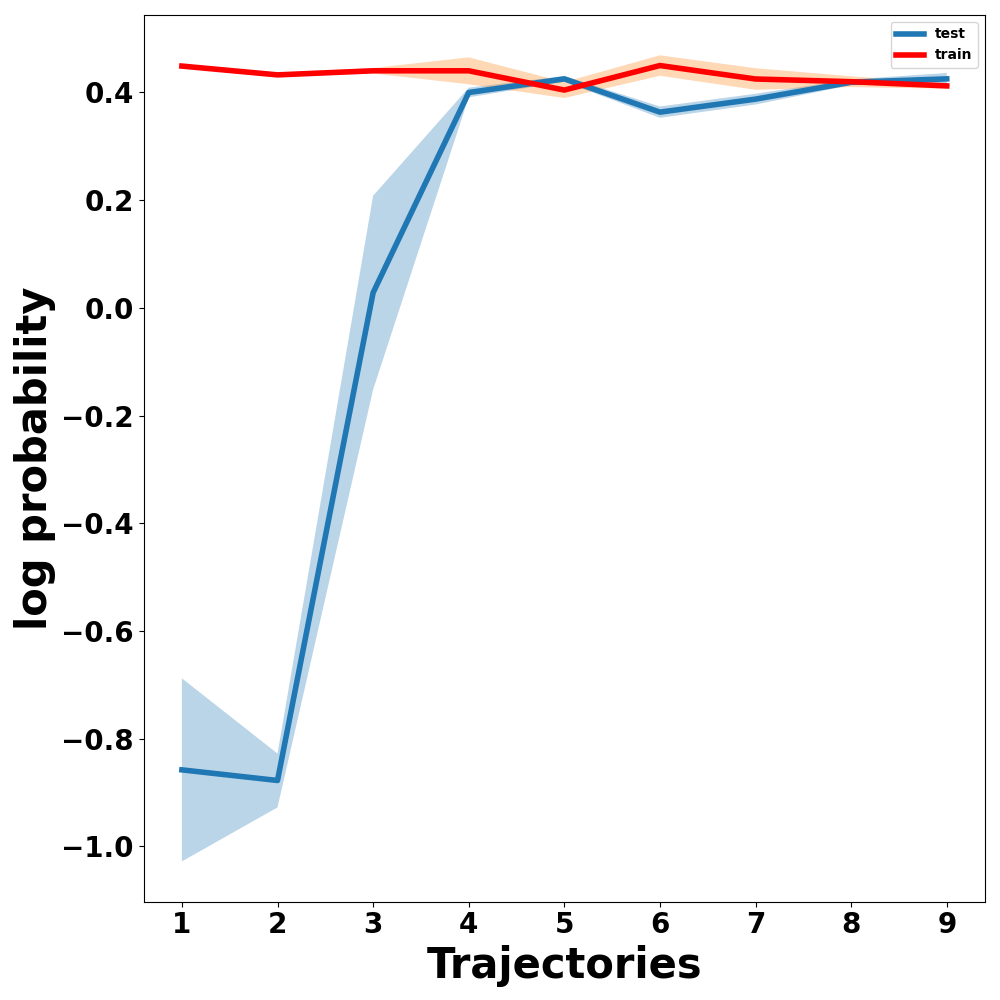}
  \end{subfigure}
  \caption{Left, learned stable dynamics for context(0.0, 0.5 and 1.); in blue, the real trajectory; in green, the generated one. The vector field represents the learned dynamic system. The middle case was not part of the training set, but properly generalized to the context. Right, obtained log-likelihoods for train and test data with different amount of training trajectories.}
  \label{fig:vitruvian_man}
\end{figure}
We have considered the problem of learning a highly nonlinear 2D limit cycles with a low dimensional context state. We have draw 10 trajectories of a man with his arms and legs in different positions, shown in Fig~\ref{fig:vitruvian_man}. The context state $\vc$ is a phase parameter that takes values between 0 and 1. 

We evaluate in this experiment the capability of our structured dynamics to generalize to both unseen position states and context state. As shown in Fig~\ref{fig:vitruvian_man}, the policy was able to impose stable motions towards our demonstrated trajectories in all the position state-space. 

To evaluate the generalization with different contexts, and given we can compute the exact log-likelihood, we computed the log-likelihood of some test trajectories in regions of the context-state that was not considered in the training. As shown, in Fig.~\ref{fig:vitruvian_man}, the policy was able to extrapolate the behavior with as few as 5 trajectories.

\subsection{Obstacle Avoidance with Stable Policies}
\begin{figure}[b]
    \centering
    \includegraphics[width=.99\textwidth]{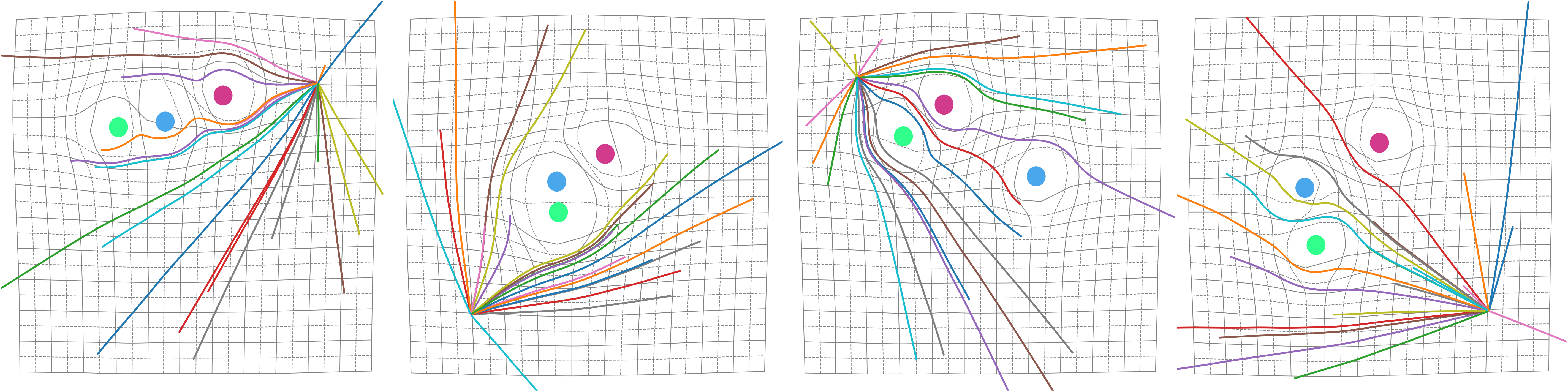}
  \caption{Learned obstacle avoidance trajectories. The deformation in the grid space represents the learned conditioned diffeomorphic mapping with the \gls{inn}. The latent linear stable dynamics are deformed to avoid the obstacles. Each colored line is a generated trajectory from a random starting pose, while the colored circles are the obstacles conditioning the \gls{inn}.}
  \label{fig:obst_avoid}
\end{figure}

Through this experiment, we wanted to validate our structured policies for learning go-to motions instead of limit cycles. We also wanted to study the capacity of our policy to contextualize in higher dimensional contexts. We considered the problem of learning an obstacle avoidance policy while being attracted to the target. We considered 3 obstacles position and the target position as the context information and learned the dynamics by \gls{mle} on the expert demonstrations.

In Fig.~\ref{fig:obst_avoid}, we show the qualitative results of the learned dynamics. The model was able to learn obstacle avoidance dynamics from expert demonstrations given arbitrary positions of the obstacles and while remaining stable towards the target goal. Figure~\ref{fig:obst_avoid} also shows the applied diffeomorphic transformation in the state-space. Given a rectangular grid in the latent space, the \gls{inn} learns a deformation of the space around the obstacle to avoid the collisions.

\subsection{Robot Pouring}
\begin{figure}[t]
\centering
\begin{subfigure}{.19\textwidth}
    \centering
    \includegraphics[width=.99\textwidth]{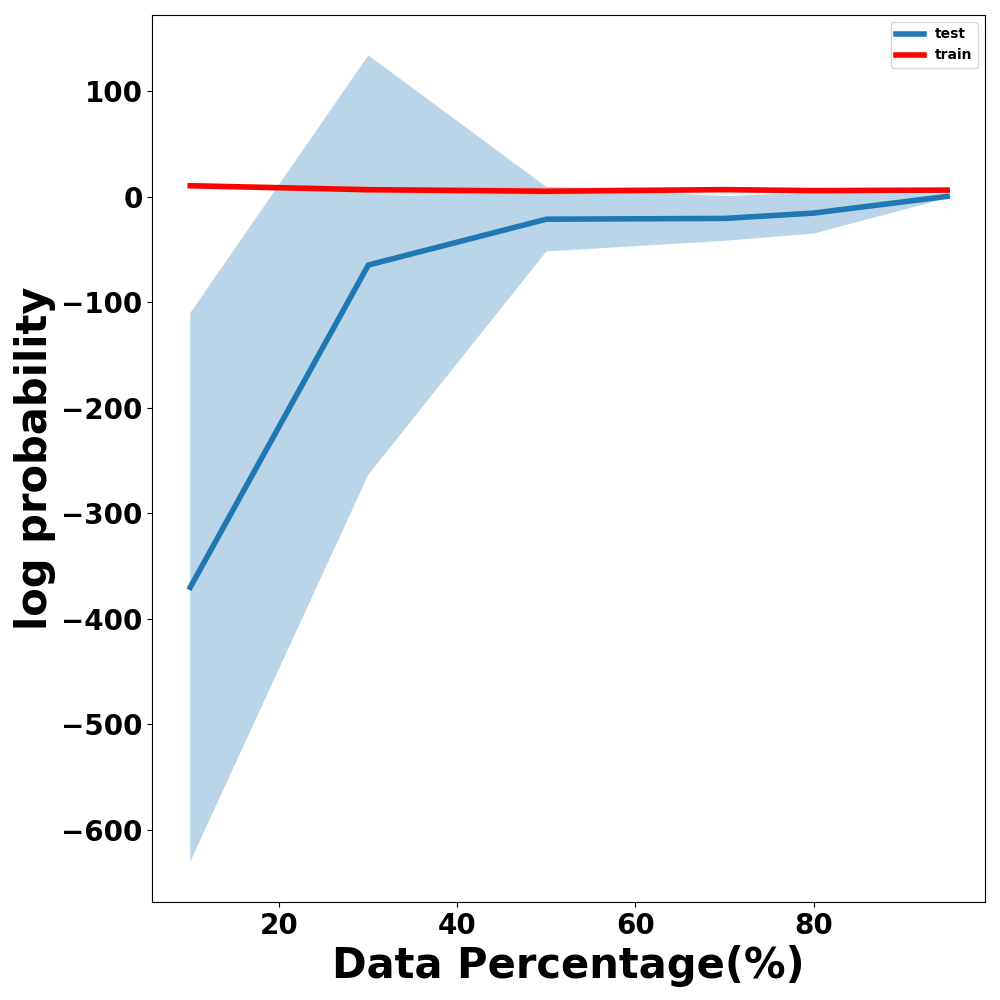}
\end{subfigure}
\begin{subfigure}{.59\textwidth}
    \centering
    \includegraphics[width=.99\textwidth]{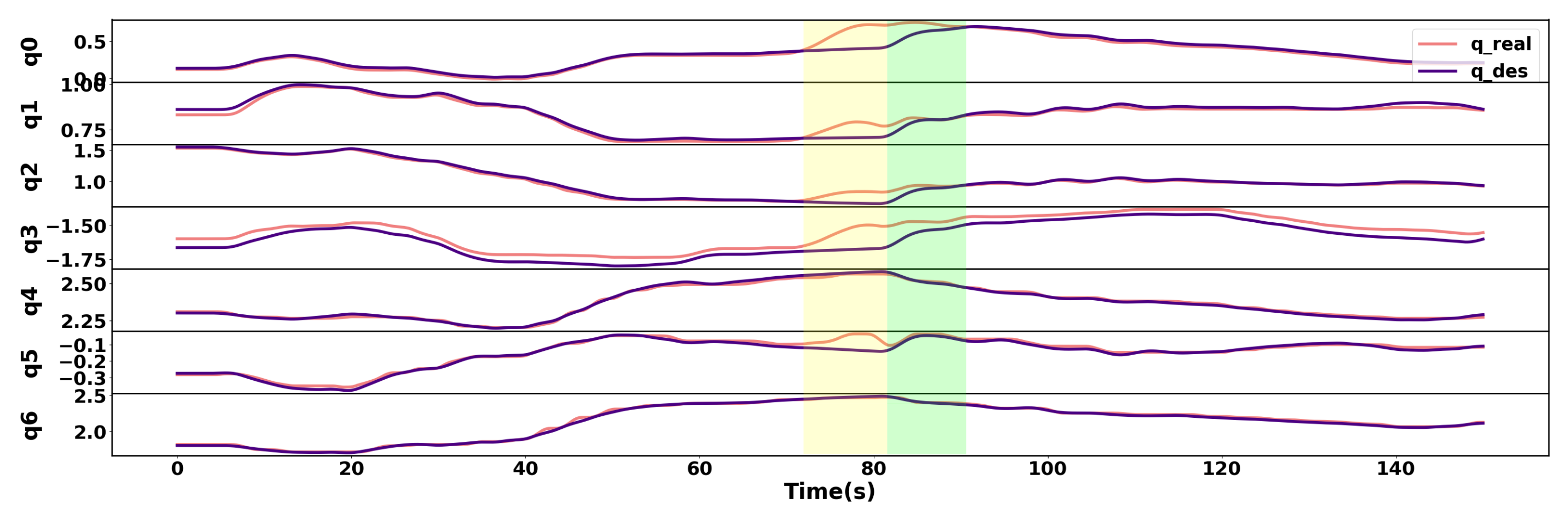}
\end{subfigure}
\begin{subfigure}{.19\textwidth}
    \centering
    \includegraphics[width=.9\textwidth]{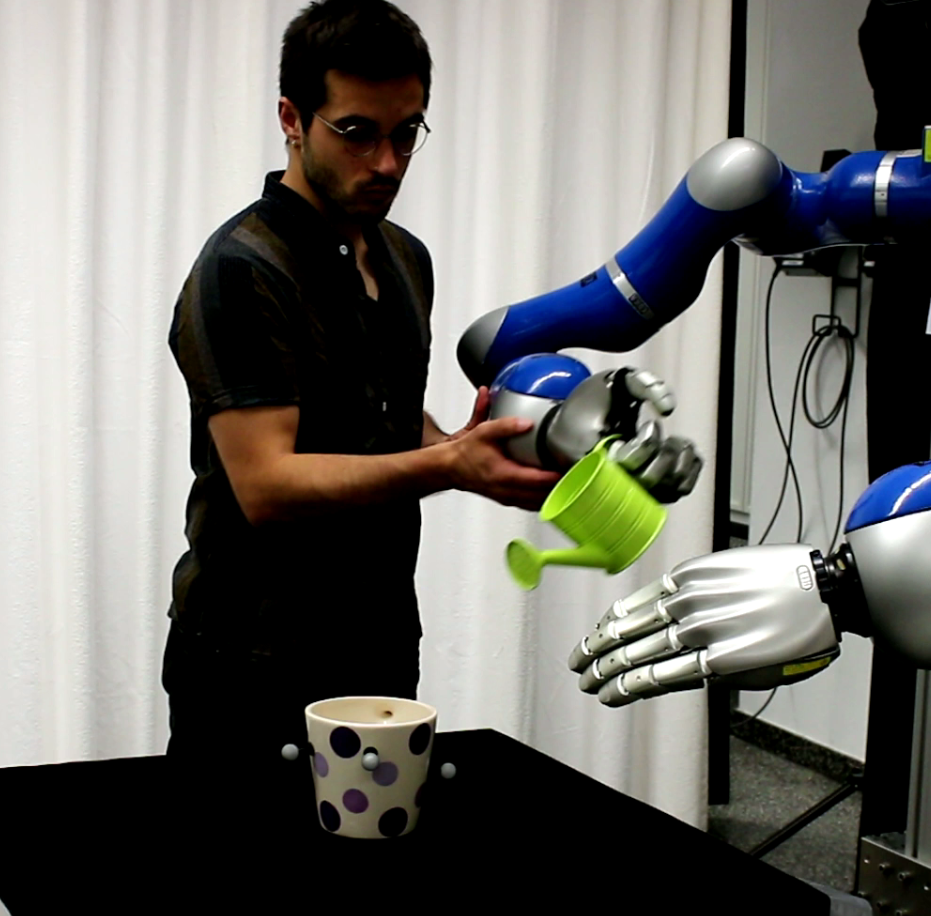}
\end{subfigure}
    \caption{Left, the obtained log probability's mean and standard deviation in the train and test data when training with different percentages of the total data. Center, recorded robot state(pink) and robot commands(purple) during pouring task. In yellow block, human disturb the robot modifying its position. In green block, our policy react to the force, generating new commands. Right, learned pouring task.}
    \label{fig:pouring_reactivity}
\end{figure}

In this experiment, we evaluate the proposed dynamics as a policy in a real robot pouring task. In this experiment, we evaluate the scalability of the proposed stable dynamics model to higher dimension dynamics. We also evaluate the robustness of the policy in front of external perturbances and context modifications. Our state is defined by the robot's joint state. The context state is the 6D(position and orientation) of the target pot. We recorded via kinesthetic teaching 220 trajectories of the desired skill with different contexts. The robot should learn the motion to water the pot without dropping the water in the trajectory towards the pot. We performed \gls{mle} in the demonstrations and as shown in Fig~\ref{fig:pouring_reactivity}(left), using the whole training data, we obtained test performances on the test data that are in line with the training ones.

We evaluate the performance of the learned dynamics in a real scenario, in which the robot's motion was perturbed both by human external forces and by the modification of the context state. In Fig.~\ref{fig:pouring_reactivity}, the center,  we show the case of adding human disturbances. In the yellow region, a human performs some external force in the robot, moving it away from the policy provided trajectory. Then the policy adapts to the disturbance generating new motion commands.

\section{Related Work}

\paragraph{Stable Dynamics} There have been several works trying to learn stable policies. One of the first approaches of this family is \gls{seds} by \cite{khansari2011learning}, which is the foundation of many other techniques~\cite{khansari2014learning,neumann2015learning,ravichandar2017learning}. Similar in spirit to our work are \cite{neumann2015learning} and \cite{perrin2016fast}. In \cite{neumann2015learning}, $\tau$-\gls{seds} is proposed. They combined a quadratic invertible transformation with \gls{seds}, increasing the representation capacity. In \cite{perrin2016fast},  the proposed invertible transformation was a locally weighted nonparametric translation. In our work, we considered a much richer family of invertible transformation such as \gls{inn} and extended the family of solutions to Stochastic Dynamic Systems, similarly to \cite{ urain2020imitation, rana2020euclideanizing}.

\paragraph{Conditioned Dynamics} Related with the context state $\vc$, most of the previous works consider simple affine transformations given the context, (\cite{schaal2006dynamic, calinon2016tutorial, khansari2014learning}) and the context information must be closely related with the desired dynamics(goal target position). In our approach, given Conditioned \gls{inn} are used as the bijective transformation, we can condition our dynamics in arbitrary contexts $\vc$ and apply highly nonlinear morphing on our dynamics, given the context.

\section{Conclusions}
We have presented a new policy representation, that can be conditioned on arbitrary context with highly nonlinear transformation while remaining globally asymptotically stable. We have presented our results in a set of Behavioural Cloning situations to prove the policy's representation power.

\medskip

\small

\clearpage

\bibliography{bibliography}  

\clearpage

\appendix

\section{Stochastic Stable Dynamics by Invertible Neural Networks}
\label{appendix:stability}

In this section, we introduce the stability proof for the model presented in \eqref{eq:iflow}. Remark, that the stability guarantees are studied in the continuous time domain, while \eqref{eq:iflow} represents a discretised version of the dynamics. We can easily obtain the discretised model by Euler-Maruyama method.

\paragraph{Model} We study the stability guarantees of a class of dynamic systems, composed by a latent($\mZ$) stable stochastic dynamic system and a deterministic, invertible transformation ($f\,: \RR^{d}\xrightarrow{}\RR^{d}$) between latent space ($\mZ$) and the observation space ($\mY$)
\begin{align}
    & d\vz(t) = v(\vz)dt + \vg(\vz) d \vB(t) \nonumber\\
    & \vy = f(\vz),
\end{align}
where  $ v : \RR^d \to \RR^d $ is a continuous function, $ \vg : \RR^d \to \RR^{d\times d}  $ is a continuous matrix function, and $ \vB : \RR \to \RR^d $ is a $d-$dimensional \emph{Brownian motion} (also called \emph{Wiener process}). We prove the stability guarantees for the dynamic system in $\mY$ space
\begin{align}
    d \vy(t) = J(\vy)v(f^{-1}(\vy))dt + J(\vy)\vg(f^{-1}(\vy))d \vB(t)
    \label{eq:observ_dyn}
\end{align}
where $J(\cdot)$ is the jacobian matrix $\frac{\partial \vy}{\partial \vz}$. 
For simplicity of the derivations, we rewrite~\eqref{eq:observ_dyn} as
\begin{align}
    d\vy(t) = v_y(\vy)dt + \vg_y(\vy) d \vB(t)
\end{align}

\paragraph{Stability Theory} We study the stability guarantees through the Lyapunov Stability method applied for Stochastic dynamics. Assume there exist a $V(\cdot,\cdot) : \RR^{d} \times \RR \xrightarrow{} \RR$, $ (\vy, t)\mapsto V(\vy, t) $ Lyapunov candidate. The time derivative for $V(\vy, t)$ can be expressed by It\^{o}'s formula
\begin{align}
    dV(\vy,t) = LV(\vy,t) dt + V_{\vy}(\vy,t) \vg_y(\vy , t) d \vB(t) \nonumber\\
    LV(\vy,t) = V_t + V_{\vy} v_y(\vy,t) + \frac{1}{2} \textrm{Tr}(\vg_y^{\intercal}(\vy , t)) V_{\vy\vy}\vg_y(\vy , t),
    \label{eq:sto_derivative}
\end{align}
where $LV(\vy,t)$ is known as the diffusion equation. 
From \eqref{eq:sto_derivative}, the expected value of $\dot{V}$ is
\begin{align}
    \mathbb{E}\left[\dot{V}(\vy,t)\right] = LV(\vy,t).
\end{align}

To prove stability of our stochastic dynamic system in $\mY$, two conditions must be satisfied. Given strictly increasing functions $\mu_1, \mu_2, \mu_3$, if
\begin{subequations}
\begin{align}
    \mu_1(|\vy|) &\leq V(\vy,t) \leq \mu_2(|\vy|), \label{eq:stLyapu1} \\
    LV(\vy,t) &\leq - \mu_3(|\vy|) \, \forall \vy \in \RR^{d}  \label{eq:stLyapu2}.
\end{align}
\end{subequations}
then, our stochastic dynamic system \eqref{eq:observ_dyn} is stochastically asymptotically stable.
\paragraph{Proof} We define the following Lyapunov candidate
\begin{align}
     V(\vy) = U(\vz) = U(f^{-1}(\vy)),
     \label{eq:lyap_candidate}
\end{align}
were $U(\cdot)$ is a valid Lyapunov candidate for the latent dynamics in $\mZ$. Given the condition \eqref{eq:stLyapu1} is satisfied for $U(\vz)$, from the equality in \eqref{eq:lyap_candidate}, the condition is also satisfied for $V(\vy)$.

To evaluate condition \eqref{eq:stLyapu2}, the diffusion equation for $V(\vy)$ must be computed
\begin{align}
       LV(\vy) = & V_{\vy} J(\vy)v(f^{-1}(\vy)) + \frac{1}{2} \textrm{Tr}((J(\vy)\vg(f^{-1}(\vy)))^{\intercal} V_{\vy\vy} (J(\vy)\vg(f^{-1}(\vy)))).
       \label{eq:lv}
\end{align}
We can rewrite  $V_{\vy}$ and $V_{\vy\vy}$ in terms of $U_{\vz}$ and $U_{\vz\vz}$
\begin{align}
    V_{\vy}(\vy) = \frac{\partial}{\partial \vy} V(\vy) = \frac{\partial}{\partial \vz} U(\vz) \frac{\partial \vz}{\partial \vy} = U_{\vz}(\vz)J^{-1}(\vy),
    \label{eq:vy}
\end{align}
where $U_{\vz}(\cdot) \,:\, \RR^{d} \xrightarrow{} \RR^{1 \times d}$. 
Finally, we can rewrite $V_{\vy\vy}$
\begin{align}
    V_{\vy\vy}(\vy) &= \frac{\partial^{2}}{\partial \vy^{2}}V(y) = \frac{\partial}{\partial \vy^{\intercal}} \Big( \frac{\partial}{\partial \vy} V(\vy) \Big) 
    = \frac{\partial}{\partial \vy^{\intercal}} \Big(U_{\vz}(\vz)J^{-1}(\vy) \Big) 
        \label{eq:vyy} \nonumber\\
    &= \frac{\partial \vz^{\intercal}}{\partial \vy^{\intercal}}\frac{\partial}{\partial \vz^{\intercal}}\Big(U_{\vz}(\vz)J^{-1}(\vy) \Big)
     J^{-\intercal}(\vy) U_{\vz\vz}(\vz) J^{-1}(\vy).
\end{align}
Rewriting \eqref{eq:lv} as
\begin{align}
    LV(\vy) = &  U_{\vz} f(\vz) +\frac{1}{2} \textrm{Tr}(\vg^{\intercal}(\vz) U_{\vz\vz} \vg(\vz)) = LU(\vz).
    \label{eq:lu}
\end{align}
By hypothesis $LU(\vz)$ satisfies the condition in \eqref{eq:stLyapu2}, therefore the condition is also satisfied by $LV(\vy)$.

The proof above shows that the dynamic system in \eqref{eq:iflow} model is globally stochastically asymptotically stable as long as the latent dynamics are globally asymptotically stable. 
In the following, we extend the stability proof for Conditioned dynamics.

\paragraph{Conditioned Model} We study the stability guarantees for a class of dynamic systems, composed by a latent($\mZ$) stable stochastic dynamic systems and a \textbf{context-based}, deterministic, invertible transformation($f\,: \RR^{d}\times \RR^{n} \xrightarrow{}\RR^{d}$), that maps $\vz \in \RR^{d}, \vc \in \RR^{n}$ to $\vy \in \RR^{d}$
\begin{align}
    & d\vz(t) = v(\vz)dt + \vg(\vz) d \vB(t) \nonumber\\
    & \vy = f(\vz, \vc),
\label{eq:CImitFlow}
\end{align}
where $\vc$ is the context variable, and $f$ is bijective between $\vy$ and $\vz$. 

\paragraph{Context Based Stability} For all $\vc \in \mC$, the proposed context-based transformation $f$ is bijective and invertible. Therefore, $\forall \, \vc \in \mC$ the conditioned model is a stable system.

However, even if for each $\vc \in \mC$ the model is stable, this property does not ensure that the Conditioned model will be stable. In order to ensure stability, we must consider additional assumptions on the context dynamics. 
Assume that the context variable evolves as follows
\begin{align}
    \dot{\vc}(t) = w(\vy(t), \vc(t))
\end{align}
with  $||\vc(t) - \vc^{*} || < \delta$ for all $t>t_0$ and $\vc(t) \rightarrow \vc^{*}$ as $t \rightarrow \infty$ i.e, the dynamics are asymptotically stable. Thus, as $t \rightarrow \infty$, the conditioned model will evolve towards the particular stochastic stable dynamic case
\begin{align}
    & d\vz(t) = v(\vz)dt + \vg(\vz) d \vB(t) \nonumber\\
    & \vy = f(\vz, \vc^{*}).
\label{eq:CImitFlow_opt}
\end{align}
Thus, the Conditioned model will be globally stochastically asymptotically stable. 

\subsection{Behavioural Cloning Learning}
We can apply \gls{mle} for behavioural cloning in our model. Given a set of trajectories $\gD_{\tau} = (\tau_0, \tau_1, \dots, \tau_K)$, where each trajectory is composed of $T$ context and robot state
$\tau_i = (\vy_0^{i}, \vc_0^{i}, \dots, \vy_T^{i}, \vc_T^{i}, )$
\begin{align}
    \theta^{*} = \arg \max_{\theta} \log p(\gD_{\tau};\theta)
\end{align}
where,
\begin{align}
    p(\gD_{\tau};\theta) = \prod_{i=0}^{K}p(\vy_0^{i}|\vc_{0}^{i};\theta)\prod_{t=1}^{T} p(\vy_{t}^{i}|\vy_{t-1}^{i},\vc_{t-1}^{i};\theta).
\end{align}
From \eqref{eq:CdynNF}, we can compute the exact probability for $p(\vy_{t}^{i}|\vy_{t-1}^{i},\vc_{t-1}^{i};\theta)$, and so, we can directly compute the \gls{mle}.

\end{document}

%% file: math_commands.tex
\usepackage{amsmath,amsfonts,bm}

\def\1{\bm{1}}

\def\RR{\mathbb{R}}

\def\va{{\bm{a}}}

\def\vc{{\bm{c}}}

\def\vg{{\bm{g}}}

\def\vq{{\bm{q}}}

\def\vs{{\bm{s}}}

\def\vy{{\bm{y}}}

\def\vz{{\bm{z}}}
\def\vB{{\bm{B}}}

\def\vtau{{\boldsymbol{\tau}}}

\def\mC{{\bm{C}}}

\def\mY{{\bm{Y}}}
\def\mZ{{\bm{Z}}}

\DeclareMathAlphabet{\mathsfit}{\encodingdefault}{\sfdefault}{m}{sl}
\SetMathAlphabet{\mathsfit}{bold}{\encodingdefault}{\sfdefault}{bx}{n}

\def\gD{{\mathcal{D}}}



%% file: acronyms.tex
\newacronym{bc}{BC}{Behavioural Cloning}
\newacronym{il}{IL}{Imitation Learning}
\newacronym{irl}{IRL}{Inverse Reinforcement Learning}
\newacronym{lfd}{LfD}{Learning from Demonstration}
\newacronym{em}{EM}{Expectation Maximization}
\newacronym{promp}{ProMP}{Probabilistic Movement Primitives}
\newacronym{dmp}{DMP}{Dynamic Movement Primitives}
\newacronym{seds}{SEDS}{Stable Estimator of Dynamical Systems}
\newacronym{gmr}{GMR}{Gaussian Mixture Regressor}
\newacronym{gpr}{GPR}{Gaussian Process Regressor}
\newacronym{lwr}{LWR}{Locally Weighted Regressor}
\newacronym{kmp}{KMP}{Kernelized Movement Primitives}
\newacronym{clf}{CLF}{Control Lyapunov Function}
\newacronym{wsaqf}{WSAQF}{Weighted Sum of Asymmetric Quadratic Function)}
\newacronym{nilc}{NILC}{Neurally Imprinted Lyapunov
Candidate}
\newacronym{clfdm}{CLF-DM}{Control Lyapunov Function-based Dynamic Movements}
\newacronym{ciflow}{C-IFlows}{Conditioned ImitationFlows}
\newacronym{iflow}{IFlows}{ImitationFlows}
\newacronym{cnmp}{CNMP}{Conditional Neural Movement Primitives}
\newacronym{tpgmm}{TP-GMM}{Task Parameterized GMM}

\newacronym{gcl}{GCL}{Guided Cost Learning}

\newacronym{mle}{MLE}{Maximun Likelihood Estimation}
\newacronym{sde}{SDE}{Stochastic Differential Equation}
\newacronym{ode}{ODE}{Ordinary Differential Equation}

\newacronym{probs}{ProbS}{Probabilistic Segmentation}
\newacronym{crf}{CRF}{Conditional Random Fields}
\newacronym{ppca}{PPCA}{Probabilistic Principal Component Analysis}
\newacronym{gmcc}{GMCC}{Generalized Multiple Correlation Coeficcient}
\newacronym{hri}{HRI}{Human-Robot Interaction}
\newacronym{ip}{IP}{Interaction Primitives}
\newacronym{hmm}{HMM}{Hidden Markov Model}
\newacronym{cac}{CAC}{Canonical Correlation Coefficient}
\newacronym{rv}{$R_v$}{$R_v$ Coefficient}
\newacronym{dcor}{dCor}{Distance Correlation}
\newacronym{dtw}{DTW}{Dynamic Time Warping}
\newacronym{edr}{EDR}{Edit Distance With Real Penalty}
\newacronym{twed}{TWED}{Time Warp Edit Distance}
\newacronym{r2}{$R^2$}{Coefficient of Determination}
\newacronym{sqp}{SQP}{Successive Quadratic Programming}
\newacronym{rkhs}{RKHS}{Reproducing Kernel Hilbert Space}
\newacronym{icnn}{ICNN}{Input-Convex Neural Network}

\newacronym{pca}{PCA}{Principal Component Analysis}

\newacronym{maf}{MAF}{Masked Autoregressive Flow}
\newacronym{iaf}{IAF}{Inverse Autoregressive Flow}
\newacronym{node}{N-ODE}{Neural ODE}
\newacronym{nsflow}{NSF}{Neural Spline Flows}
\newacronym{cnf}{CNF}{Conditional Normalizing Flows}
\newacronym{ffjord}{FFJORD}{Free-form Jacobian of Reversible Dynamics}

\newacronym{gan}{GAN}{Generative Adversarial Networks}
\newacronym{vae}{VAE}{Variational Autoencoders}
\newacronym{nf}{NF}{Normalizing Flows}
\newacronym{inn}{INN}{Invertible Neural Networks}